# How Far Should Self-Driving Cars 'See'? Effect of Observation Range on Vehicle Self-Localization

Mahdi Javanmardi, Ehsan Javanmardi, and Shunsuke Kamijo, *Member, IEEE*

*Abstract*— Accuracy and time efficiency are two essential requirements for the self-localization of autonomous vehicles. While the observation range considered for simultaneous localization and mapping (SLAM) has a significant effect on both accuracy and computation time, its effect is not well investigated in the literature. In this paper, we will answer the question "How far should a driverless car observe during self-localization?" We introduce a framework to dynamically define the observation range for localization to meet the accuracy requirement for autonomous driving, while keeping the computation time low. To model the effect of scanning range on the localization accuracy for every point on the map, several map factors were employed. The capability of the proposed framework was verified using field data, demonstrating that it is able to improve the average matching time from 142.2 ms to 39.3 ms while keeping the localization accuracy around 8.1 cm.

## I. INTRODUCTION

Self-localization is one of the fundamental pillars of autonomous driving, allowing self-driving vehicles to obtain their accurate position online and, based on that position, perform other essential tasks such as environment perception, path planning, and navigation. It has been shown that autonomous vehicle localization in urban environments requires an accuracy exceeding what can be achieved by GPS-based positioning systems [1]. Simultaneous Localization and Mapping (SLAM) has been a popular alternative for robots and autonomous vehicles for more than two decades. Among different SLAM techniques, localization in a previously built map is a viable approach [2] and has been employed practically in several autonomous driving prototypes by companies such as Google [3] and Toyota [4].

The first step for localization using a previously built map is to roughly identify where the vehicle is located (e.g. using GPS/IMU). When this step is accomplished, the local observation from the vehicle is registered to the previously built map using data association algorithms. With the advent of light detection and ranging (LiDAR), it is considered as a fundamental sensor technology for self-driving cars which provides high-resolution, three-dimensional observation from vehicle surroundings [5]–[7]. Because of its accuracy and direct 3D measurement, LiDAR is very popular for mapping and localization of autonomous vehicles.

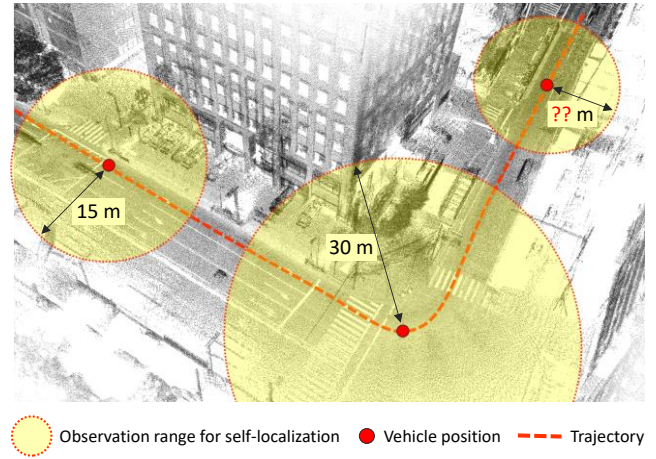

Figure 1. Concept of dynamically defining observation ranges for vehicle self-localization.

To optimize for safety and provide the vehicle the time it needs to quickly react to changing road conditions or unexpected events, autonomous vehicles need to see as far ahead as possible (e.g. 200 m ~ 300 m) [8]. Similarly, increasing the observation range in the self-localization gives the vehicle more features which helps improving the localization accuracy. However, the longer the scanning range considered, the larger the input points will become, which in turn linearly increases the computation time of the registration.

It is apparent that, for the best performance of autonomous vehicle self-localization, both accuracy and computation time are equally important. To ensure the safe operation of the vehicle it is considered that the localization accuracy should be below a threshold (usually around 20 cm [9]) at all times and the algorithm should meet real-time constraints. There are dozens of efforts to improve the accuracy or execution time of LiDAR-based localization or both. Some researchers focused on the registration algorithm and data association problem. For example, Fontanelli *et al.* [10] proposed a simple and fast RANSAC-based registration algorithm for on-line and accurate localization. Others utilized new landmarks for registration such as traffic signs [11] to improve accuracy. In [12], the authors achieved localization accuracy in the 10cm-range by an efficient compression of laser data using probabilistic maps. In [13], the authors propose a registration algorithm that employs a 2D Gaussian mixture map to speed up the registration process. As shown, different approaches in the literature try to solve the accuracy and computation time problem of the vehicle localization. Nevertheless, to the best of our

Mahdi Javanmardi is with the Institute of Industrial Science, the University of Tokyo, Tokyo, Japan (phone: +81-3-5452-6273; fax: +81-3-5452-6274; e-mail: mahdi@kmj.iis.u-tokyo.ac.jp).

Ehsan Javanmardi is with the Institute of Industrial Science, the University of Tokyo, Tokyo, Japan (e-mail: ehsan@kmj.iis.u-tokyo.ac.jp).

Shunsuke Kamijo is with the Institute of Industrial Science, the University of Tokyo, Tokyo, Japan (e-mail: kamijo@iis.u-tokyo.ac.jp).

knowledge, there is not a work focusing on the observation range considered during the localization.

In our previous work [14], a framework to model localization error within a given map based on map factors was proposed. In [15], this framework was utilized for adaptive resolution refinement of normal distributions transform (NDT) map. In this paper, we will investigate the effect of observation range on both accuracy and execution time of the vehicle self-localization (Section II), and propose an idea that, for every point within the map, the observation range can be defined adaptively based on the surrounding environment and the level of localization error (Section III). The framework proposed in [14] is then utilized to model the localization error along an experiment path for different observation ranges (Section IV) to apply the method to real application. Finally, the modeled error is used to define the least observation range which can provide the vehicle with the required accuracy for the autonomous driving (Section V). These observation ranges can be calculated off-line and embedded into the map for on-line localization. Figure 1 shows the general idea of this research.

## II. Effect of Observation Range on Self-Localization

Before explaining the relation between observation range and self-localization performance, we will briefly outline different steps of the range sensor-based self-localization using a priori-built map. As a case study, we pick one popular registration algorithm which is the normal distributions transform (NDT) [16]. In general, we can divide the self-localization into three parts: pre-built map, local observation of the vehicle, and a matching algorithm to register the observation to the map. In this paper, our focus is on the local observation of the vehicle. In NDT, before each registration, the input observation is filtered and subsampled using a regular grid of a certain resolution to speed up computations and to avoid bias from uneven point distribution [16]. This subsampling is one of the parameters which affects overall performance. After the filtering and subsampling, registration is started from an initial guess which can be calculated using GPS, IMU, and previous position of the vehicle. For registration, local hill-climbing from the initial pose is used.

Two necessary metrics for self-localization are accuracy and computation time. As mentioned in the introduction, the considered range for environment observation can affect these two metrics. Therefore, the observation range is an important parameter for self-localization which, in fact, is not well-studied in the literature. To the best of our knowledge, this is the first study to report the effect of the observation range on the self-localization performance. Increasing the observation range provides more features (e.g. curbs, building edge, road signs, etc.) which helps improve the localization accuracy. At the same time, by seeing a longer range, the number of input points will become larger, which increases the computation time. Therefore, to choose an optimum observation range, the trade-off between accuracy and computation time should be considered.

It is apparent that to keep the computation time as low as possible we want to employ the minimum possible observation range. As the distribution of features around the vehicle differs from place to place, the minimum observation range to meet a certain accuracy cannot be static. To evaluate the effect of observation range on both accuracy and computation time of the vehicle localization, we conducted an urban experiment in Tokyo, Japan.

### A. Study area

The experiment was conducted near Shinjuku, a dense urban area of central Tokyo, Japan. The experiment path is 1.1 km and included narrow streets, wide streets, multiple crossings, buildings, different road facilities and trees which is shown in Figure 2a.

### B. Experiment method

Our experiment vehicle was equipped with two multilayer laser scanners (Velodyne LiDAR VLP-16), one single layer laser scanner (SICK LMS511), GPS, IMU, and CAN (Figure 2b). The experiment consisted of two separate phases: map reconstruction and localization test.

During the map reconstruction phase, laser scanner No. 1 (installed horizontally) was used for localization of the vehicle and scanner No. 2 (installed vertically) was used for mapping. We made sure that the vehicle was driven very slowly (e.g. less than 3.6 km/h), so that scan distortion was kept very small and negligible in the mapping process. In addition, collected reflective luminance by scanner No. 3 was used to improve the map accuracy using the method proposed in [17]. Figure 3 shows a part of the generated map.

For the localization phase, we used the laser scanner mounted on the roof (scanner No. 1) and registered the observation to a priori-built map using NDT technique.

For every point on the experiment route, the localization performance was evaluated considering different observation ranges (i.e. 10 m to 50 m with an interval of 5 m). For each point, we generated a set of initial offsets for the localization around the ground-truth position, evaluating how accurately and how quickly NDT can perform the localization (Figure 4). The same process was repeated for each observation

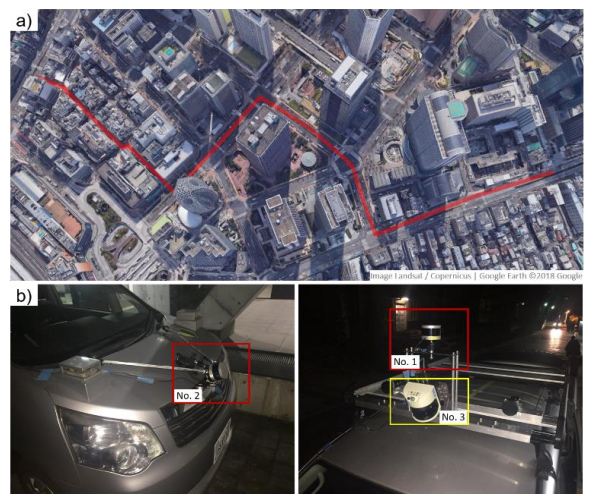

Figure 2. a) Study area near Shinjuku, Tokyo. b) Experimental platform equipped with LiDAR, GPS, IMU and CAN.

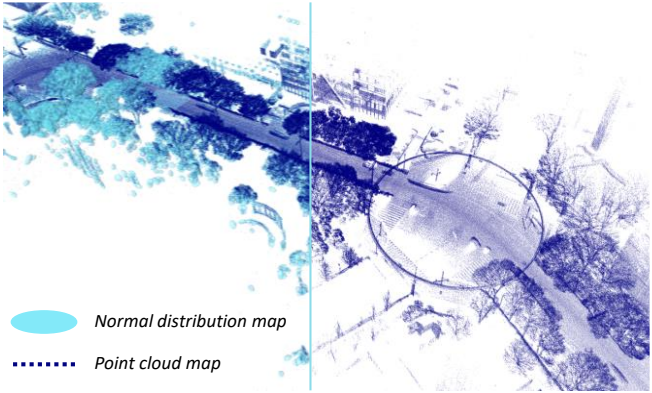

Figure 3. A part of the generated point cloud map and NDT map of the study area.

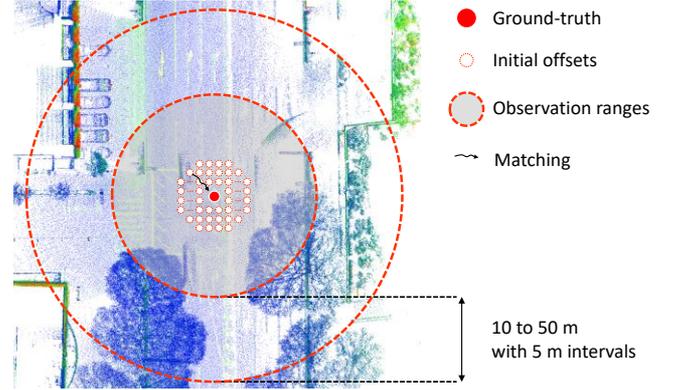

Figure 4. Error calculation for every point on the experiment path.

range and mean error and computation time were reported.

All the computations were conducted on an off-the-shelf PC with an 8-core, 3.50 GHz Intel Xeon E3-1270 V2 CPU and 16 GB of RAM running the 64-bit version of the Ubuntu 16.04 operating system.

### C. Experimental Result

Localization result using the maximum range of the laser scanner, 100 m, was considered as ground truth for the evaluation. Figure 5 demonstrates the mean localization error along the experimental route for different observation ranges. As can be seen in the figure, the scanning range and localization error have an inverse relationship: as the observation range increases, the margin of error decreases. For ranges above 50 m, the localization accuracy is almost the same as the maximum range. Figure 6 shows the average and maximum matching time for different observation ranges. As shown in the figure, the matching time increases as expected by increasing the considered range for the localization. Table I shows the reported mean absolute 3D error and matching time for different observation ranges.

### D. Discussion

The main parameter that affects the matching time is the number of points within the input observation. In NDT matching, the input scan is first subsampled and then the matching is performed. The experimental results show that the number of points in each scan sequence after the subsampling has a direct relationship with the execution time of the matching. This means that if the observation range is decreased, it will decrease the number of points within the range, which will effectively decrease the matching time. Also, it can be understood from Figure 5 that the localization error differs from point to point. If the localization accuracy is high enough, maybe an acceptable performance can be achieved even with a shorter range.

TABLE I. ERROR AND MATCHING TIME FOR DIFFERENT RANGES

| range (m) | 10 | 15 | 20 | 25 | 30 | 35 | 40 | 45 | 50 |
|---|---|---|---|---|---|---|---|---|---|
| error (cm) | 15 | 15 | 8.6 | 6.0 | 4.1 | 2.6 | 1.7 | 0.9 | 0.3 |
| μ (ms) | 13 | 31 | 47 | 64 | 101 | 117 | 128 | 136 | 142 |
| max (ms) | 24 | 54 | 80 | 104 | 163 | 187 | 205 | 215 | 224 |

### III. DYNAMIC OBSERVATION RANGE: IDEAL CASE

As highlighted in the previous section, changing the observation range during localization results in a trade-off between the accuracy and computation time. In the literature, we can find some discussions about the required localization accuracy for autonomous driving. While there is no agreement on a specific number, decimeter-level accuracy is considered as a common understanding among the researchers [2], [9]. In general, for real-time applications such as autonomous driving, every event (i.e. self-localization) should response within specified time constraints. However, it is apparent that finishing the tasks as early as possible makes the system more robust and provides

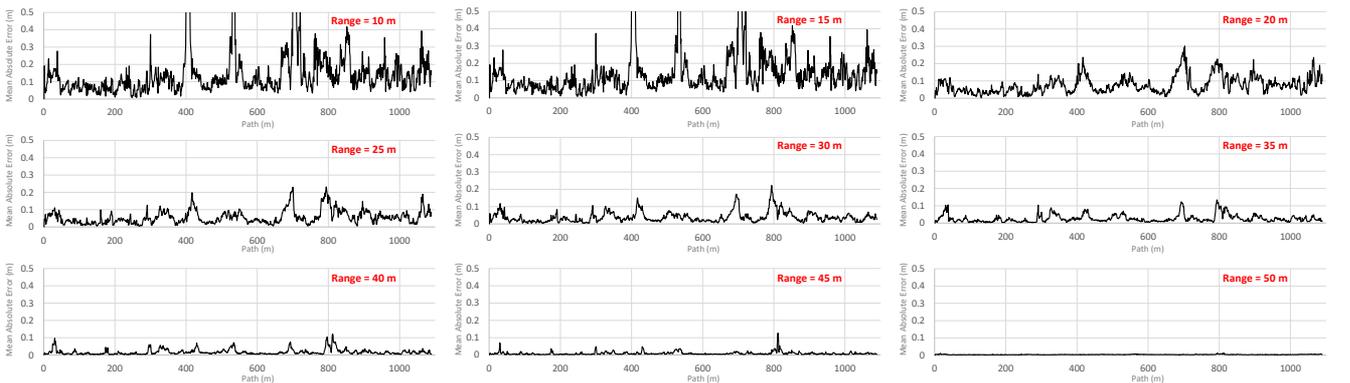

Figure 5. Localization error for different observation ranges.

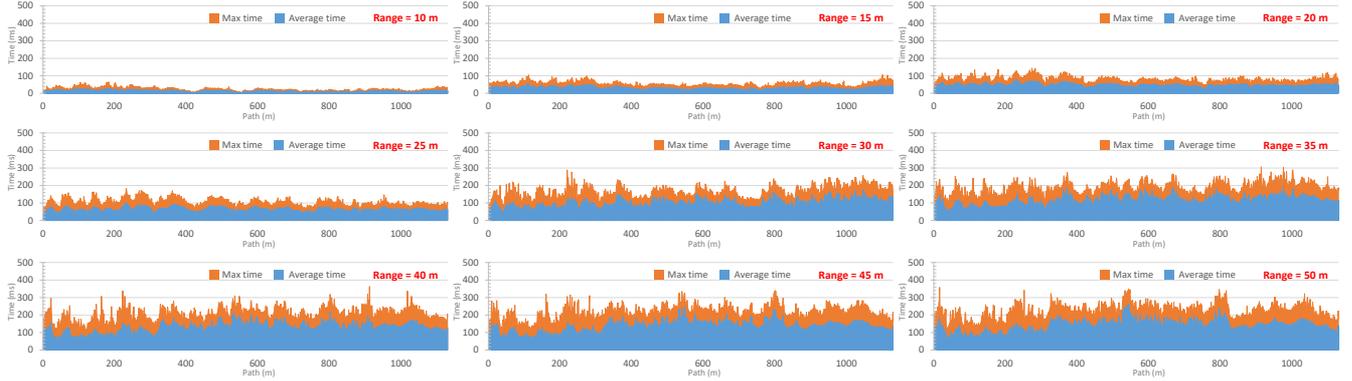

Figure 6. Average and maximum matching time for different observation ranges.

more time for other tasks such as perception and navigation.

In an ideal case, considering that we know the localization error for different observation ranges for every point on the path, we can choose the shortest range which meets the accuracy requirement. Figure 7 shows the shortest observation ranges along the path, considering two different localization error requirements, 10 cm, and 25 cm.

By dynamically choosing the range, average matching time for the experimental route can be improved from 142.2 ms to 16.4 ms which is almost one order of magnitude faster. Comparison of the average matching time is shown in Figure 8. Using dynamic observation range (DOR) could also reduce the maximum matching time along the path from 261.0 ms to 54.9 ms. TABLE II compares the matching time for the entire path for static scanning range of 50 m, DOR of 10 cm error and DOR of 25 cm error.

TABLE II. MATCHING TIME EVALUATION FOR DOR(IDEAL CASE)

|  | Static range (50 m range) | Dynamic range (10 cm error) | Dynamic range (25 cm error) |
|---|---|---|---|
| μ (ms) | 142.2 | 41.7 | 16.4 |
| max (ms) | 261.0 | 156.8 | 54.9 |

## IV. SCANNING *RANGE* ESTIMATION FRAMEWORK

In the ideal DOR, it was assumed that the localization errors for different observation ranges are given for the entire path. However, it is not a realistic assumption, and that is why defining dynamic ranges which meet the accuracy requirement is not an easy task.

Localization error of a certain location is related to the way that features are distributed around the vehicle. Thereby, observation ranges should be chosen based on the distribution of features within local vicinity of the vehicle.

In [14], we have proposed a way to model localization error by only looking at the map. In this paper, different factors proposed in our previous work is utilized to model the localization error for different observation ranges. Some of these factors are:

- $feature\_count$
- $dimension\_count$ and $dimension\_ratio$
- $occupancy\_ratio$
- $normal\_entropy$
- $r\_average$
- $score\_entropy$

which will be briefly explained later. As we use NDT technique for the localization, the aforementioned metrics are calculated based on normal distributions (NDs) of the map within local vicinities (Figure 9).

### A. $feature\_count$

In order to evaluate the sufficiency of features within a local vicinity, $feature\_count$ is used which shows the number of NDs within defined ranges. It is apparent that increasing observation range will lead to an increase in $feature\_count$.

### B. $dimension\_count$ and $dimension\_ratio$

In order to evaluate the quality of features, they are

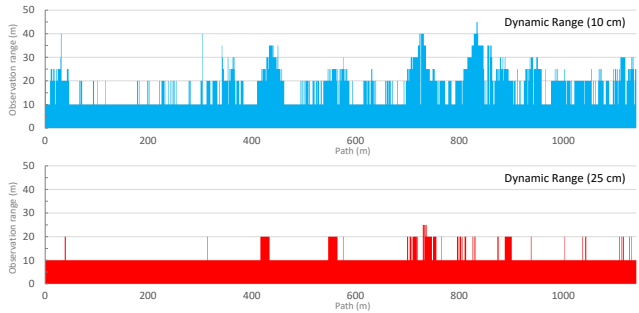

Figure 7. Shortest observation ranges for 10 and 25 cm error (ideal case).

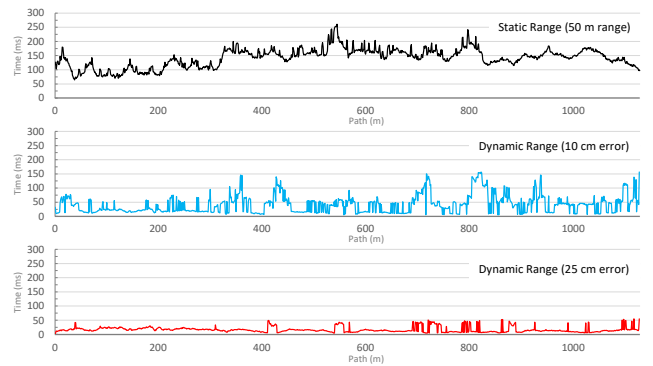

Figure 8. Matching time for static range and dynamic range localization.

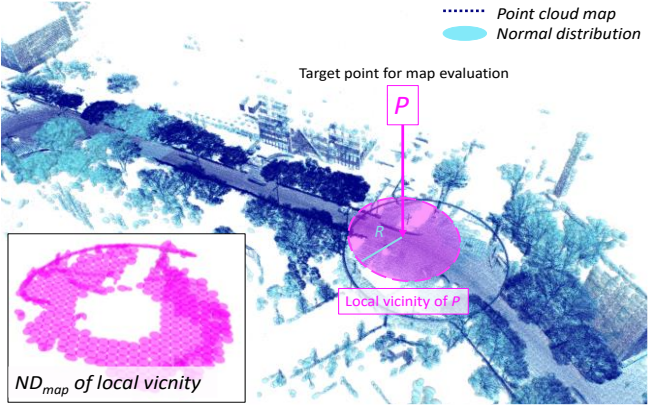

Figure 9. Local vicinity arount point *P* with observation range of *R*.

divided into three categories based on their Eigenvalues and three dimension behaviors which are defined as:

$$a_{1D} = \frac{\sigma_1 - \sigma_2}{\sigma_1}, a_{2D} = \frac{\sigma_2 - \sigma_3}{\sigma_1}, a_{3D} = \frac{\sigma_3}{\sigma_1}, \quad (1)$$

where $\sigma_i$ is standard deviation for $i_{th}$ Eigenvalue, $a_{nD}$ is the $n_{th}$ dimension behavior of a feature. If $a_{1D} \gg a_{2D}, a_{3D}$, the feature is considered to be $1D$ feature, if $a_{2D} \gg a_{1D}, a_{3D}$, the feature is a $2D$ feature, and finally if $a_{3D} \gg a_{1D}, a_{2D}$, then the feature is $3D$ feature. Using this definition, the $feature\_count$ is divided into $D1\_count$, $D2\_count$, and $D3\_count$. In addition, the ratio of each feature over all features is considered as a new factor: $D1\_ratio$, $D2\_ratio$, and $D3\_ratio$.

### C. occupancy_ratio

$occupancy\_ratio$ shows how much the surrounding environment is occupied with the features. This factor considers the feature sufficiency from the vehicle view.

In order to calculate this factor, the local vicinity is converted to a depth image. $occupancy\_ratio$ is the ratio of the occupied cell over all cells of the depth image and it is calculated simply as follows:

$$occupancy_{ratio} = \frac{occupied\_cell}{all\_cells} \quad (2)$$

### D. normal_entropy

$normal\_entropy$ estimates the diversity of features' normal direction in local vicinity which useful for evaluating the layout of the features.

$normal\_entropy$ is calculated based on a histogram of the direction of features' normal as:

$$H_{normal} = -\sum_{i=1}^{b} P_{normal}(i)(log_2 P_{normal}(i)), \quad (3)$$

where b is the number of bins, and $P_{normal}(i)$ is the probability of occurrence of $i_{th}$ bin.

If the normal of the features within the local vicinity has a higher degree of dispersity, the entropy of the normal angle histogram will be higher. On the other hand, a lower $normal\_entropy$ means that features within the local vicinity have similar orientations which can affect the localization.

Therefore, localization accuracy and $normal\_entropy$ are in a direct relationship.

### E. r_average

$r\_average$ shows the average distance of the features from the vehicle and is calculated as

$$r_{average} = \sum_{i=1}^{m} r_i, \quad (4)$$

where $m$ is the number of features in the local vicinity and $r_i$ is the distance to the $i_{th}$ feature from the vehicle.

### F. score_entropy

To formulate similarity within a local vicinity defined by different ranges, $score\_entropy$ is used. $score\_entropy$ can be used as a tool to find repeating patterns in the surrounding environment within a certain range.

The score of registering a local vicinity of a specific range to itself with a shift $\vec{v}$ is calculated as

$$s(\vec{v}) = -\sum_{k=1}^{n} \tilde{p}(T(\vec{v}, x_i)), \quad (5)$$

where $\tilde{p}$ is a simplified log-likelihood function of the nearest ND to the point $x_i$ in the range and $T(\vec{v}, x_i)$ is the transformation function which transforms $x_i$ with the 2D transformation vector $\vec{v}(x, y)$. In order to calculate $score\_entropy$, first the score is converted to probability distribution based on a specific range for the shift $\vec{v}$ and its entropy is calculated.

### G. Modeling error for different ranges

The aforementioned factors are used to estimate localization error along the experiment path for different observation ranges using Random Forest Regression.

## V. DYNAMIC OBSERVATION RANGE: EXPERIMENT

The modeled localization error for different observation ranges was used to define the required observation range for self-localization and answer the question "How far should self-driving cars 'see'?" In order to define the required observation range for every point on the experiment route, localization error was modeled for different observation ranges between 10 m to 50 m with an interval of 5 m for every point on the path. Based on these errors, the observation ranges were defined dynamically to keep the error under 10 cm. Figure 10 shows the scanning ranges defined by this method and compares their computation times with the static range method. As a result, we could reduce the average matching time from 142.2 ms to 39.3 ms while keeping the mean error around 8.1 cm. The average range used for the experimental road was 16.9 m which is much shorter than the original 50 m. For 81.5% of the experiment path, the defined range could keep the localization error within 10 cm. Figure 11 illustrates the observation ranges on an aerial image of the area. As can be seen in the figure, the selected observation ranges fit the surrounding environments well.

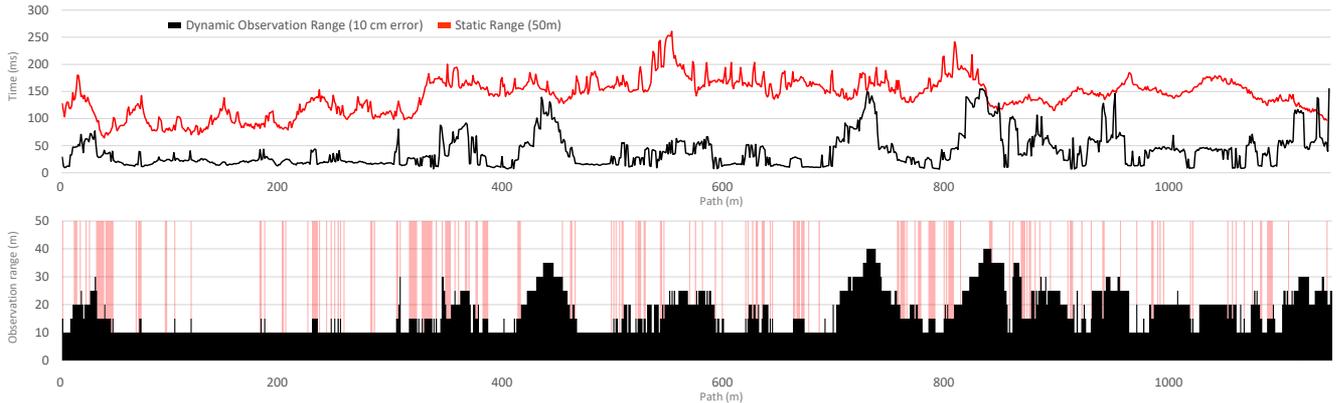

Figure 10. top) Comparison of matching between Static Range (50 m) and Dynamic Observation Range Localization considering 10 cm error, down) the selected observation ranges to keep 10 cm accuracy. The pink bars indicate the locations where the selected range could not keep the localization error within 10 cm. This happens when the modeled error for the selected observation range was less than 10 cm while the real error was exeeding 10 cm.

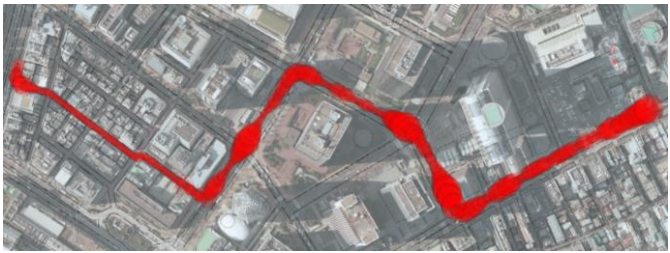

Figure 11. Dynamic observation ranges for the experiment area.

## VI. Conclusion and Future Work

In this paper, the effect of observation range on the vehicle localization performance was demonstrated. Based on this effect, a localization method which defines the observation range dynamically based on the localization accuracy was proposed. Finally, to model the effect of observation range on localization accuracy, the error modeling framework using several map factors were utilized and the dynamic range localization was performed based on the error model. The results from Shinjuku, Tokyo demonstrated the capability of the proposed localization method by reducing the average matching time from 142.2 ms to 39.3 ms while keeping the localization accuracy around 8.1 cm. In the future, we will use a larger data set to evaluate the capability of the dynamic range localization more deeply.

## Acknowledgment

The authors would like to thank Kelvin Wong, Institute of Industrial Science, the University of Tokyo, for constructive criticism of the manuscript.